\documentclass[10pt,conference,a4paper]{IEEEtran}
\IEEEoverridecommandlockouts
\usepackage{cite}
\usepackage{amsmath,amssymb,amsfonts}
\usepackage{algorithmic}
\usepackage{graphicx}
\usepackage{textcomp}
\usepackage{multirow}
\usepackage{xcolor}
\usepackage{url}
\def\BibTeX{{\rm B\kern-.05em{\sc i\kern-.025em b}\kern-.08em
    T\kern-.1667em\lower.7ex\hbox{E}\kern-.125emX}}
\begin{document}

\title{Assessing the Severity of Health States based on Social Media Posts}
\author{
\IEEEauthorblockN{Shweta Yadav}
\IEEEauthorblockA{\textit{Department of Computer Science} \\
\textit{Wright State University}\\
Dayton, Ohio, USA \\
shweta@knoesis.org}
\and
\IEEEauthorblockN{Joy Prakash Sain}
\IEEEauthorblockA{\textit{Department of Computer Science} \\
\textit{Wright State University}\\
Dayton, Ohio, USA \\
sain.9@wright.edu}
\and
\IEEEauthorblockN{Amit Sheth}
\IEEEauthorblockA{\textit{Artificial Intelligence Institute} \\
\textit{University of South Carolina}\\
Columbia, South Carolina, USA \\
amit@sc.edu}
\and
\centering
% \begin{center}
\phantom{jkfbbbbbbbbbbffffffffffcdsbaciejw}
\IEEEauthorblockN{
Asif Ekbal\IEEEauthorrefmark{1}, Sriparna Saha\IEEEauthorrefmark{2} and Pushpak Bhattacharyya\IEEEauthorrefmark{3}}

\IEEEauthorblockA{
\hspace{55mm} Department of Computer Science \& Engineering \\
\hspace{55mm} Indian Institute of Technology Patna\\
\hspace{55mm} Bihar, India\\
\hspace{55mm} Email: \IEEEauthorrefmark{1}asif@iitp.ac.in,
\IEEEauthorrefmark{2}sriparna@iitp.ac.in,
\IEEEauthorrefmark{3}pb@iitp.ac.in,
}
% \end{center}
}

% \IEEEauthorblockN{Asif Ekbal}
% \IEEEauthorblockA{\textit{Department of CSE} \\
% \textit{Indian Institute of Technology Patna}\\
% Patna, India \\
% asif@iitp.ac.in}
% \and
% \IEEEauthorblockN{Sriparna Saha}
% \IEEEauthorblockA{\textit{Department of CSE} \\
% \textit{Indian Institute of Technology Patna}\\
% Patna, India \\
% sriparna@iitp.ac.in}
% \and
% \IEEEauthorblockN{Pushpak Bhattacharyya}
% \IEEEauthorblockA{\textit{Department of CSE} \\
% \textit{Indian Institute of Technology Patna}\\
% Patna, India \\
% pb@iitp.ac.in}
% }

\maketitle

\begin{abstract}
The unprecedented growth of Internet users has resulted in an abundance of unstructured information on social media including health forums, where patients request health-related information or opinions from other users. Previous studies have shown that online peer support has limited effectiveness without expert intervention. Therefore, a system capable of assessing the severity of health state from the patients' social media posts can help health professionals (HP) in prioritizing the user’s post. In this study, we inspect the efficacy of different aspects of Natural Language Understanding (NLU) to identify the severity of the user’s health state in relation to two perspectives(tasks) (a) Medical Condition (i.e., Recover, Exist, Deteriorate, Other) and (b) Medication (i.e., Effective, Ineffective, Serious Adverse Effect, Other) in online health communities. We propose a multiview learning framework that models both the textual content as well as contextual-information to assess the severity of the user’s health state. Specifically, our model utilizes the NLU views such as sentiment, emotions, personality, and use of figurative language to extract the contextual information. The diverse NLU views demonstrate its effectiveness on both the tasks and as well as on the individual disease to assess a user’s health\footnote{The manuscript is accepted for publication at 25th International Conference on Pattern Recognition.}.
\end{abstract}


\section{Introduction}
The volume of patient-generated healthcare data is experiencing an immense growth.
% According to IBM, $2.5$ quintillion bytes of healthcare data are generated globally every day. 
The primary contributors to this enormous amount of data are social networks, forums, and blogs where patients share their medical problems and treatment experiences, including adverse reactions to medical products. According to Pew Internet \& American Life Project \cite{fox2011social,yadav-etal-2019-unified,yadav-etal-2018-multi,yadav-etal-2018-medical}, almost 80\% of Internet users in the US explore health-related topics in online health forums. Among them, 63\% look for information about specific medical problems, and nearly 47\% look for medical treatments or procedures. 

In the online health communities (OHC) and support groups, healthcare professional (HPs) provide clinical intervention when required, for example, when patients need any interpretation or details of clinical concepts or medical consultation. In these settings, patients benefit from the knowledge of both peer-patients' and HPs simultaneously \cite{huh2013text,ieee-is,ieee-cim}.
\cite{kummervold2002social} reports that the users on health forums expect the involvement of HPs for a quality suggestion and virtual observation. Another survey carried out by Pew Internet Research\footnote{{https://pewrsr.ch/2wEL6JU}}, shows that more than 80\% of the patients preferred to consult the HPs rather than peer-patients for the information on prescription drugs, medical diagnoses, and alternative treatment options.
However, the participation of HPs in the large-scale discussion forums is time-consuming. A study \cite{huh2013patient} on the $6$ most active communities on \url{WebMD.com} data showed the level of HPs  participation was observed to be extremely low (only $~4.7$\% of the posts were answered).
Hence, novel strategies are necessary for prioritization of the blog-posts based on the severity of ones' health states that could assist the HPs to efficiently select the posts that need their expertise for making an effective and timely response. Specifically, we explore two important facets of the health state as proposed by \cite{yadav2018medical} that can help in assessing the severity of ones' health states based on their social media posts discussed as follows:
\begin{itemize}
\item \textbf{Status of the medical/health condition} (e.g., a patient whose medical condition is  deteriorating even after clinical trials tends to be more severe than the patient who has started experiencing the medical symptoms.) 
\item \textbf{Consequences of the medication/treatment} (e.g., a patient reporting an adverse drug effect will be more severe than a patient whose treatment was ineffective.)
\end{itemize}
In Table-\ref{ex-usecase1}, we provide the examples and description of the classes associated with the above described facets.
\begin{table*}[]
\centering
\resizebox{\textwidth}{!}{
\begin{tabular}{l|l|l|l|l}
\hline
\multicolumn{2}{c|}{\textbf{Task 1: Medical Condition}} & \multicolumn{3}{c}{\textbf{Task 2: Medication}} \\ \hline
\multicolumn{1}{c|}{
\textbf{Health Blog-post}} & \textbf{Class-labels} & 
\multicolumn{1}{c|}{
\textbf{Health Blog-post}} & \multicolumn{2}{l}{\textbf{Class-labels}} \\ \hline
\textit{\begin{tabular}[c]{@{}l@{}}``my high resolution CT scan came back\\ normal...I'm doing better after a long, breathless\\ blue journey"\end{tabular}} & Recover & \textit{\begin{tabular}[c]{@{}l@{}}``I think the plaquenil is helping- been on it \\ for almost 3 months"\end{tabular}} & \multicolumn{2}{l}{Effective} \\ \hline
\textit{\begin{tabular}[c]{@{}l@{}}``Been having eye problems ...lots of swelling\\ redness and eye discharge."\end{tabular}} & Exist & \textit{\begin{tabular}[c]{@{}l@{}}``I have been on Hizentra for almost a year...I \\ don't seem to be getting sick as much or as\\ bad (yay) but things areen't normal, that's for\\ sure"\end{tabular}} & \multicolumn{2}{l}{Ineffective} \\ \hline
\textit{\begin{tabular}[c]{@{}l@{}}``It's been just over three months and I'm \\ actually feeling worse! my IgG levels are rising \\ from 766 to 1423, but I don't feel good..."\end{tabular}} & Deteriorate & \textit{\begin{tabular}[c]{@{}l@{}}``I was given propranolo for migraine \\ associated vertigo. I've only taken 5 mgs fir \\ last 2 nights and I've had bad nausea since \\ starting it, I want to know if I can just stop \\ taking it now without any problems "\end{tabular}} & \multicolumn{2}{l}{Serious Adverse Effect} \\ \hline
\textit{\begin{tabular}[c]{@{}l@{}}``Not everyone uses or likes Facebook. Let's \\ remember there are so many who are looking for \\ information and support on this cvid forum.-- \\ Our voice can help Other"\end{tabular}} & Other & \textit{\begin{tabular}[c]{@{}l@{}}``Hello ladies, I am curious if anyone has any \\ knowledge or experience with kidney \\ symptoms resulting from using either \\ Gammagard or other IG therapy brands"\end{tabular}} & \multicolumn{2}{l}{Other} \\ \hline
\end{tabular}
}
\caption{Examples and description of classes for the Task 1: Medical Condition and Task 2: Medication. The `Exist' class belongs to those users' who are reporting their medical problem (symptoms) but are not under any form of treatment/medication. The `Deteriorate' class consist of those users' whose medical problem has worsen even after taking the treatment/medication.}
\label{ex-usecase1}
\end{table*}
% A system capable of gleaning such insights from social media text can have other several usage such as:\\
% \begin{itemize}
% \item \textbf{Use-Case 1: Personalized Health Assistance:} The analysis of users' self-reported text can help in making inferences about their mental and physical states. These inferences can further be utilized to develop online solutions to direct the users to health information, assistance and also to provide personalized interventions.
% \item \textbf{Use-Case 2: Pharmacovigilance:} The quantity and near-instantaneous nature of social media provide potential opportunities for real-time monitoring of Adverse Drug Reaction \cite{sarker2015utilizing}. Furthermore, the pharmaceutical industries can utilize social media to monitor patients’ or healthcare professionals’ opinion towards marketed medicines \cite{chou2009social}.
% \end{itemize}
% \begin{figure}[h]
% \centering
% \includegraphics[width=\textwidth]{Severity_UseCases-1&2.png}
% \caption{Exemplar description of use-cases. The top figure describes use case 1 and down figure is the description of use case 2.}
% \label{ex-usecase}
% \end{figure}

To evaluate our study, we have used the benchmark dataset made available through the LRE map \cite{yadav2018medical}. The dataset utilized the blog-posts from two popular medical discussion forums, namely \url{`patient.info'} and \url{`dailystrength.org'} over four groups, namely asthma, allergy, depression and anxiety.  The prior studies \cite{yadav2018medical,yadav2018multi} have treated this task as a traditional document classification tasks and propose the neural network framework which is able to capture the content level information from the user's blog-post. However, the social-media texts often carry slang terminology, grammatical errors, figurative languages and hold in the information which is highly contextual. In such situations, mining only linguistic information turns out to be generally inefficient and urges for extra information/clues. The study conducted by \cite{wallace2014humans} also demonstrates this requirement by showing how traditional classifier fails in instances where humans need additional context. They further illustrate the importance of speaker and topical information associated with the text to incorporate such context.\\
\indent Advancement in NLU technology, is one of the most promising avenues for discovering vital contextual information from such data. Motivated by that, in this study we hypothesize that the NLU views such as emotions, sentiments, personality and usage of figurative language can help in understanding the vital contextual clues and discourse of the blog post required for detecting the health states. Ahead in this section, we have provide the detailed motivation behind utilizing the various NLU views in the multi-view learning framework.\\
\indent The emotions that patients express towards their personal situation could be an important indicator to understand their health.
Understanding the emotion can capture a user’s mental and physical health and can be applied to microblog posts \cite{wang2012harnessing} for behavioral decision making. Similarly, in mental health diagnosis, certain personality traits correlate with the diagnosis. Personality can be defined as characteristics patterns of an individual's thinking, behaving, and emotional feeling. Hence, automatic identification of an individual's personality type can have a wide range of applications in personalized health diagnosis \cite{cortellese2009personality} and to discover the user's behavior. \\
\indent We also explore the sentiments, which focus on extracting opinions and affects from the textual content \cite{pang2008opinion}, it seems natural that incorporating these knowledge can be helpful to discover emotional statement in the area of online health text classification.
Previous research has provided evidence to suggest that people’s mental and physical health can be predicted by analyzing their sentiment \cite{yadav2018medical,carrillo2018feature} and the word they use \cite{gottschalk1969measurement,stiles1992describing}.\\
% Social media texts offer some inherently distinct challenges in the form of figurative language (FL). FL include metaphoric, sarcastic and ironic scripts that often emerge in the vicinity of informal human communication. In medical forums, patients seeking support for their medical problems, often use sarcasm to express their emotions (e.g., ``Lol Im just a big ball of anxiety fun.”). 
\indent Similarly, in sentiment analysis, the presence of figurative language (FL) such as sarcasm in a text can work as an unexpected polarity reverser, which may undermine the accuracy of the system, if not addressed adequately \cite{bosco2013developing}. \\
\indent In this work, we propose a multi-view learning framework \cite{kumar2020edarkfind} that jointly models both the content and the contextual-information specific for the tasks. It begins by processing contextual information using several NLU views such as emotions, personality, sentiments, and use of figurative language (FL).
Following the contextual modeling phase, we perform the content modeling using the Bidirectional Encoder Representations from Transformers (BERT) to extract the task-agnostic view. The task-agnostic view is then fused with the various NLU derived views to obtain the final representation used for predicting the health of patients.  
% Motivated by these facts, we develop a deep neural network based framework that utilizes the textual content (derived from Convolutional Neural Network (CNN)) as well as several NLU features to assess a user’s health state from the social media posts.\\
% We hypothesize that the NLU can help in understanding the vital contextual clues and discourse of the blog post required for detecting the health states which motivated us to augment the NLU information in the system.
Our contributions include:
\begin{enumerate}
\item A multi-view learning framework that allows the integration of different semantics captured from social media texts to support several aspects of natural language understanding (NLU) required by the health severity assessment task.
\item A comprehensive evaluation of the framework by testing it on a publicly available dataset and comparing the performance against the state-of-the-art baselines.
\item A demonstration of the effectiveness of various views such as emotion, sarcasm, personality, and sentiment on the diseases/disorders which provides the complementary information for assessing the severity of health states.
\end{enumerate}

\section{Related Work}
Literature shows increased attention on the OHCs for computationally discovering patient health \cite{davidson2019evolution}. A majority of these studies follows a qualitative approach based on the manual categorization of posts by the domain experts. The categorization includes: (i) the type of support \cite{blank2010differences,coulson2007social}, (ii) the type of emotion and sentiment expressed \cite{ruthven2018isolated}, and (iii) discussion on other illness specific topics or adverse drug effects \cite{huber2011decision,yadav-etal-2019-unified}. Below, we describe some of the prior research that utilized social media text of patients.\\
\indent Several techniques have been devised to automatically process the OHCs content, uncover the user behaviours\cite{huh2013patient} and characteristics \cite{huh2014weaving}. \cite{10.1093/jamia/ocw093} in their study formulated the OHCs from a social support viewpoint and defined three variables : type of support, source of support, and setting in which the support is exchanged on online cancer communities. \cite{lu2017understanding} developed a new content analysis method to (a) recognize various healthcare participants, (b) discover currents trends, and (c) analyze the sentiments expressed by different healthcare participants from lung cancer, diabetes, and breast cancer forums. \cite{info:doi/10.2196/jmir.6834} proposed a text mining technique to classify the user's participation based on the different types of social support, such as informational support, emotional support, and companionship. Further, they developed a supervised machine learning approach to predict whether user will be churn from OHC. 
The challenge defined in \cite{milne2016clpsych} aimed to automatically classify the user posts from an online mental health forum into four different categories (crisis/red/amber/green) according to need of urgent attention. \cite{huh2013text} developed a text classification technique for assisting the moderators in OHCs. Specifically, they devised a classification scheme to automatically categorize the blog-post over the codes such as: `Asking for medical information', `Asking for peer-patients' experience', `General chatting', and `Miscellaneous', using WebMD's online diabetes community data. Some of the other prominent research in this area includes work of \cite{o2018suddenly,balyan2019using}. Recently, \cite{yadav2018medical} has introduced the novel annotation scheme for analyzing medical sentiment on social media text that can capture the severity of the user's health states. They utilized a CNN to understand the possible sentiment. \cite{yadav2018multi} further extended the study utilizing a multitask learning framework to capture the multiple facets of medical sentiments simultaneously.

\section{Materials and methods}
First, we define the task and detail the approach used in the study. Later, we describe the dataset and provide the training details, followed by the baseline models.
\subsection{Task Definition:}
For a given medical forum post $M$, consisting of $n$ sentences, i.e., $M = \{s_1, s_2, s_3,.....s_n\}$, the task is to predict the two aspects of health status `$Y$' \& `$Z$' from a discrete set of medical conditions `$Y$=\{\textit{Recover, Exist, Deteriorate, Other}\}' and medications `$Z$=\{\textit{Effective, Ineffective, Serious Adverse Effect, Other}\}'. 

\subsection{Summary of the Approach}
\begin{figure*}[h]
\centering
\includegraphics[width=\textwidth,height=9cm]{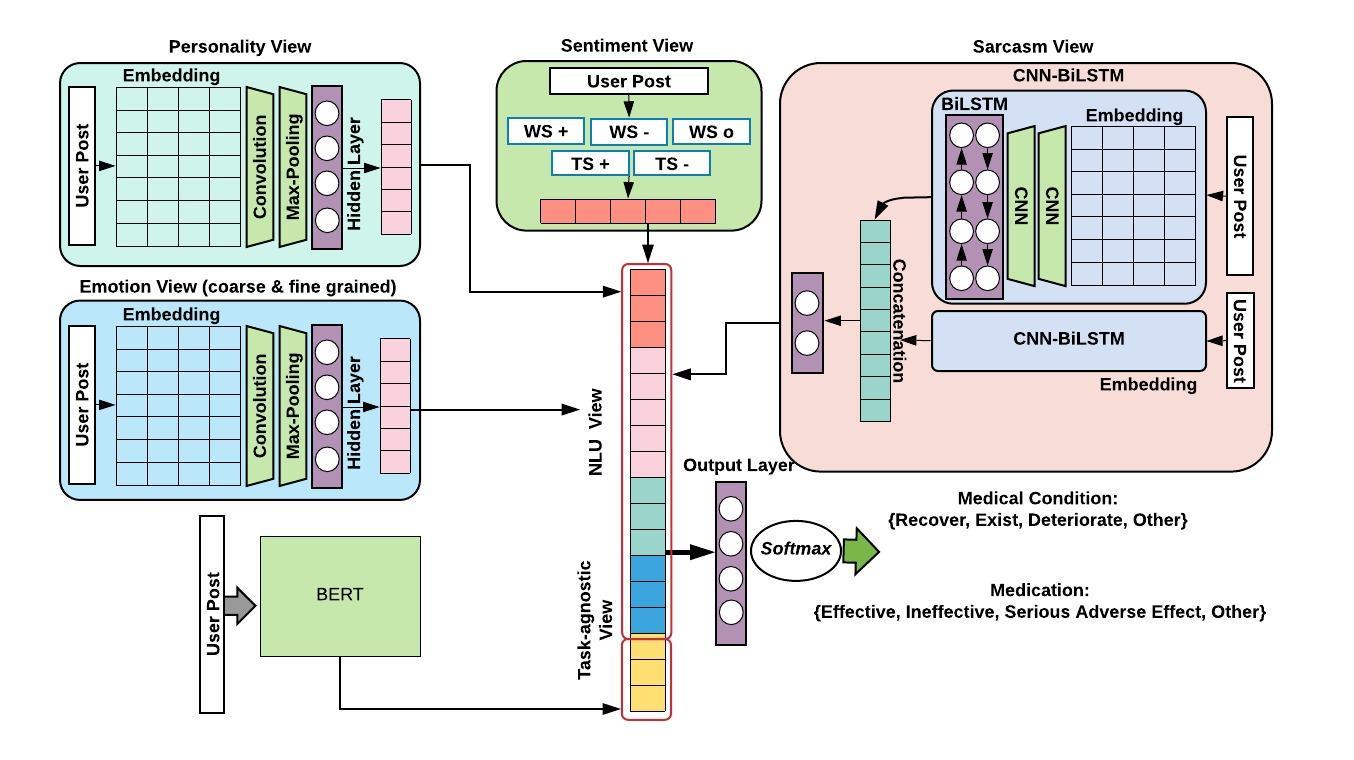}
\caption{Proposed architecture for predicting the health severity assessment task.}
\label{fig-model}
\end{figure*}

Given the user's forum post $M$ to be classified, the proposed system leverages both task-agnostics and context specific views from the text. In order to capture task-agnostic view, we model $M$ using BERT model to obtain a vector representation of the medical blog post. BERT generate the abstract representation of words by capturing bi-directional context in the input sentence \cite{jacovi2018understanding}.  Basically, BERT aims to learn the deep bidirectional representations from the Transformer stack \cite{vaswani2017attention}. 
% BERT also provides the text representation by modelling a special $[CLS]$ token in the architecture. The learn encoder representation associated with the $[CLS]$ token is able to capture the semantic of the whole text. We use this representation as the task-specific feature for the medical forum post.} 
For modeling the contextual views, the proposed system utilizes the information of multiple NLU based features/views (emotion, sarcasm, personality, and sentiment) extracted from the user forum post. For extracting emotion, sarcasm and personality views we adopt the domain adaptation approach \cite{glorot2011domain} where we train a model in one domain (i.e., publicly available gold data), and extract the features on another domain (i.e., our dataset). Finally, we fused the NLU views with the BERT generated task-agnostic views which are used to categorize the medical forum post. Figure-\ref{fig-model} shows the architecture of our approach.

\subsection{Task-agnostic View} \label{content-features}
We generated the task-agnostic views as follows:
% We obtained the first view\footnote{In this paper view and embedding are interchangeably used.} from the pre-trained language representation networks.
The task-agnostic view are generated using the BERT network. We employ the pre-trained BERT model\footnote{\url{shorturl.at/nDJPY}} having $12$ Transformer layers ($L$), each having $12$ heads for self-attention and hidden dimension of $768$, to extract the feature representation of the medical forum post. The pre-trained model has shown the state-of-the-art performance in various natural language processing tasks \cite{maillard2019jointly,hewitt2019structural}.  The pre-trained BERT model is highly efficient in generating the task-agnostic input representation from the transformer architecture \cite{Vaswani2017AttentionIA}. This enables even the low-resource tasks to benefit from deep bi-directional architectures \cite{devlin2018bert} and the unsupervised training framework to obtain the pre-trained network.
% Generic word embeddings such as word2vec \cite{mikolov2013distributed}, which is pre-trained on huge corpus, have demonstrated remarkable success when used as features for supervised learning problems. Sentence level embeddings perform better than simple word-level transfer \cite{cer2018universal}. Hence, we take two state-of-the-art pre-trained embedding, namely Universal Sentence Encoder and BERT and extract the above features. Even \cite{devlin2018bert}, one of the state-of-the-art models for Natural Language Processing, allows one to create word-level and sentence level embedding. 
We perform extensive experiment and to obtain the effective representation of the medical forum post representation. Our experimental studies conclude that the aggregating the last three layers of the BERT model achieves the best result in our experiment.
% We noticed that the last four layers produce the best representation of the sentence. 
Given a forum post $M$ consisting of $n$ tokens $\{t_1, t_2, t_3, \ldots, t_n\}$. We use the Word Piece tokenizer \cite{wu2016google} to tokenize the sentence. We use the special $[CLS]$ token representation as the task-specific feature for the medical forum post $M$. We represent this task-agnostic view as the content feature $\hat{d}_{cont}$.

\subsection{Contextual (NLU) View}
In this section, we will discuss in details all the generated NLU (emotion, sarcasm, personality, and sentiment) views:
\subsubsection{\textbf{Emotion View}}
Humans are emotional beings; emotion carries an intrinsic role in human life. 
It influences our decision making \cite{van2010interpersonal}, shapes our behavior \cite{baumeister2007emotion}, and affects mental and physical health \cite{extremera2006emotional}.\\
\indent Towards this, we study the user’s forum post on five primary emotions such as \textit{`anger', `disgust', `joy', `fear',} and \textit{`sadness'} that can assist the model in capturing the overall health condition of an individual. We also examine the post on more fine-grained emotions (i.e., \textit{`valence', `arousal',} and \textit{`dominance'}) that reveals the user’s state of feeling \cite{mehrabian1974approach}.\\
\indent In order to extract the emotion views from forum post, we leverage a system similar to the CNN network \cite{kim2014convolutional}. We utilize the benchmark emotion EmoBank-2017 dataset \cite{buechel2017emobank} and EmoInt-2017 dataset \cite{mohammad2017wassa} to train the model for fine and coarse-grained emotion analysis, respectively. After training the model, we extract the feature representation for each forum post.\\
\indent Given a forum post $M$ consisting of $n$ words, after extracting the pooled representation, we pass it to a hidden layer to generate the view representation as follows:
\begin{equation}
% \scriptsize
F_{e}=\varphi(W_e \hat{d}_{e} + b_e)
\end{equation}
where $\varphi$ is a non-linear activation function, $W_e$, and $b_e$ are the weight, and bias of the last hidden layer. The pooled feature with the given window size $k$ is generated as follows:
% \begin{dmath}
\begin{equation}
\begin{aligned}
d_{e} & = pool(conv(emb(M), k;\theta)))
\end{aligned}
\end{equation}
% \end{dmath}
where $emb(.)$, $conv(.)$ and $pool()$ are the embedding, convolution and pooling operations similar to standard CNN model. $\theta$ is the model parameter. The final extracted feature is $\hat{d}_{e}$ is obtained by concatenating the multiple features of different window size.
The fine and coarse grained emotion views $F_e^f$ and $F_e^c$ are extracted using the aforementioned equation.

% %%%%%%%%%%%%%%%%%%%%%%%%%%%%%%%%%%%%%%%%%%
\subsubsection{\textbf{Sarcasm View}}
Humans seamlessly perform high-level semantic tasks by subconsciously utilizing a vast collection of composite linguistic units along with their background knowledge to visualize the reality. Social media texts often contain FL. The presence of FL makes it challenging to process for any NLP applications, sentiment analysis in particular.
In sentiment analysis, the presence of FL such as sarcasm in a text can work as an unexpected polarity reverser, which may undermine the accuracy of the system, if not addressed adequately \cite{bosco2013developing}.
In medical forums, patients seeking support for their medical problems, often use sarcasm to express their emotion.\\
\indent In the following example, one patient with anxiety problem describes her health condition sarcastically, writing as:\\
\textit{``... It goes from pain to slight discomfort.. I cant move. Great way to start the day !"}\\
\indent Here, the phrase ``\textit{Great way to start the day !}'' is presented in the sarcastic sense to express the medical condition of not feeling good.
To model this feature for each forum post in our framework, we utilize the approach proposed by \cite{ghosh2017magnets}.
The model utilizes 2-layers of CNN followed by a Long Short Term Memory (LSTM) \cite{lstm1} network. We compute the sarcastic scores as follows: 
\begin{equation}
% \scriptsize
F_{s}=\sigma( W_s.{V}_{c} + b_s)
\end{equation}
where $\sigma$ represents the \textit{sigmoid} activation function. ${V}_{c}$, $W_s$ and $b_s$ are the input, weight, and bias term of the output layer. The input vector of the output layer $V_c$ is the concatenated feature vector that combines the features extracted by the LSTM layers, which capture the state of mind of a user.  
\subsubsection{\textbf{Personality View}}
One's behavior characterizes personality, sympathy, emotion, thought process, and motivation. Our personality impacts many preferences in our lives such as decision making, life choices, physical and mental health.
For extracting personality view, we employ an approach similar to the one used for emotion view. We utilize the corpus curated by \cite{matthews1999personality}, which contains $2,400$ essays labeled with big-five personality traits, to train the model. We utilize the CNN as a learning model for the multi-label personality detection task. 
After the training, the CNN model is used to infer the personality traits present in each medical forum post by extracting the activation of the CNN’s last hidden layer vector which we call as the \textit{personality vector}.  After the training, the model is used for extracting the feature representation.
The extracted personality vector can be represented as:

\begin{equation}
% \scriptsize
F_{p}=\varphi(W_p \hat{d}_{p} + b_p)
\end{equation}
where $\varphi$ is a non-linear activation function, and $\hat{d}_{p}$ (i.e., extracted pooled feature, $W_p$, and $b_p$ are the input, weight, and bias of the last hidden layer.\\
\subsubsection{\textbf{Sentiment View}}
We generate the sentiment views from the forum posts as described below:\\
% \begin{enumerate}
% \item 
\textbf{1) }\textbf{Word-level Sentiment (WS)}: Sentiment clue words provide important features in deciding the sentiment of the users. Besides, the inclusion of negation to the sentiment word can change the polarity. For example, there is positive sentiment in ``I'm \textbf{stable}" but after including negation like ``I'm \textbf{not stable}", the sentiment polarity changes. Briefly, there are two types of sentiment events by which we can capture the sentiments of users: occurrences of sentiment words (SW), occurrences of sentiment words with negation (NSW).
This feature calculates the positive ($SW_{(+)}$), negative ($SW_{(-)}$) and objective ($SW_{(O)}$) score for each word by capturing the sentiment event \cite{dang2009machine}. Publicly available SentiWordNet (SWN)\footnote{\url{http://sentiwordnet.isti.cnr.it/}} is used to calculate the score for each word as follows:
\begin{equation}
% \scriptsize
\begin{aligned}
SW_+= &(tf(SW)*f_{+}(SW)\\
&+tf(NSW)*f_{-}(SW))*idf(SW)
\end{aligned}
\end{equation}
\begin{table*}[]
\centering
\resizebox{\textwidth}{!}{
\begin{tabular}{c|c|c|c|c|c|c|c}
\hline
\begin{tabular}[c]{@{}c@{}}Dataset\\ Categories\end{tabular} & \multicolumn{4}{c|}{\begin{tabular}[c]{@{}c@{}}Classes\\ (\# of instances)\end{tabular}} & \# blog-post & \# avg sentences/blog & \# avg words/blog \\ \hline
Medical Condition & \begin{tabular}[c]{@{}c@{}}Exist\\ (2,396)\end{tabular} & \begin{tabular}[c]{@{}c@{}}Recover\\ (703)\end{tabular} & \begin{tabular}[c]{@{}c@{}}Deteriorate\\ (2,089)\end{tabular} & \begin{tabular}[c]{@{}c@{}}Other\\ (432)\end{tabular} & 5,621 & 10 & 194 \\ \hline
Medication & \begin{tabular}[c]{@{}c@{}}Effective\\ (462)\end{tabular} & \begin{tabular}[c]{@{}c@{}}Ineffective\\ (613)\end{tabular} & \begin{tabular}[c]{@{}c@{}}Serious Adverse Effect\\ (1,226)\end{tabular} & \begin{tabular}[c]{@{}c@{}}Other\\ (432)\end{tabular} & 2,734 & 9 & 177 \\ \hline \hline
\end{tabular}%
}
\caption{Dataset statistics for both the categories of medical sentiment}
\label{dataset-class1}
\end{table*}
\begin{equation}
% \scriptsize
\begin{aligned}
SW_-=&(tf(SW)*f_{-}(SW)\\
&+tf(NSW)*f_{+}(SW))*idf(SW)
\end{aligned}
\end{equation}

\begin{equation}
% \scriptsize
SW_O=tfidf(SW)*f_{O}(SW)
\end{equation}
\begin{table*}[]
\centering
\resizebox{\textwidth}{!}{
\begin{tabular}{c|c|ccc|ccc}
\hline
\multirow{2}{*}{\textbf{Models}} & \multirow{2}{*}{\textbf{Techniques Used}} & \multicolumn{3}{c|}{\textbf{Medical Condition}} & \multicolumn{3}{c}{\textbf{Medications}} \\ \cline{3-8} 
 &  & \textbf{Precision} & \textbf{Recall} & \textbf{F-Score} & \textbf{Precision} & \textbf{Recall} & \textbf{F-Score} \\ \hline
 \hline
Baseline 1 & \begin{tabular}[c]{@{}c@{}}BERT\end{tabular} & 72.70 & 73.15 & 72.89 & 86.64 & 87.55 & 86.81 \\ \hline
Baseline 2 & BioBERT & 72.42 & 72.30 & 72.28 & 86.68 & 86.97 & 86.76 \\ \hline
Baseline 3 & MTL \cite{yadav2018multi} &66.71  &64.33  &65.5  &85.33  &81.90  &83.58  \\ \hline
Proposed Approach & NLU based Multi-view Learning  & 75.52 & 80.25 & 77.45 & 89.52 & 89.91 & 89.57 \\ \hline
\end{tabular}
}
\caption{Performance comparison our proposed model with the baselines on both the datasets}
\label{main-results}
\end{table*}

\begin{table}[]
\centering
 \resizebox{\linewidth}{!}{%
\begin{tabular}{c|c|c|c}
\hline
\textbf{Index} & \textbf{View} & \textbf{Medical Condition} & \textbf{Medications} \\ \hline \hline
(1) & All & 77.45 & 89.57 \\ \hline
(2) & - Emotion (coarse) & 75.08 & 87.32 \\ 
(3) & - Emotion (fine) & 75.44 & 88.61 \\ 
(4) & - Sarcasm & 77.10 & 86.82  \\ 
(5) & - Personality  & 74.91 & 87.66\\ 
(6) & - Word-level Sentiment & 74.54  & 85.94 \\ 
(7) & - Target-specific Sentiment & 75.85 & 85.27 \\ \hline 
\end{tabular}%
}
\caption{Feature ablation study: The system performance (in F1 score) by removing one view at a time }
\label{feature-ablationl}
\end{table}
Here, \textit{tf} and \textit{idf} represent the term and inverse document frequencies, respectively. $f_{+}(SW)$, $f_{-}(SW)$ and $f_{O}(SW)$ are positive, negative and objective scores, respectively obtained from the SentiWordNet. The word-level sentiment feature of a forum post having $n$ words is obtained as follows: 
% \\$WS_+= \frac{\sum_{SW \in n} SW_+}{n}$ ; $WS_-= \frac{\sum_{SW \in n} SW_+}{n}$ ; $WS_O= \frac{\sum_{SW \in n} SW_O}{n}$.     

\begin{equation}
% \scriptsize
\begin{aligned}
WS_+= \frac{\sum_{SW \in n} SW_+}{n}
\end{aligned}
\end{equation}
\begin{equation}
% \scriptsize
\begin{aligned}
WS_-= \frac{\sum_{SW \in n} SW_+}{n}
\end{aligned}
\end{equation}
\begin{equation}
% \scriptsize
WS_O= \frac{\sum_{SW \in n} SW_O}{n}
\end{equation}
\textbf{2)} \textbf{Target-specific Sentiment (TS)}:
After analyzing the validation data, we observe that approximately $92$\% of the posts depict sentiments in the context of a certain stative verbs such as `\textit{feel}', `\textit{suffer}', `\textit{experience}'. We design this feature by considering a context window of [-$5$,$5$] words and selecting the most effective stative verb. After that, negative and positive densities of a post are calculated by the frequency of the clue words to the number of words in the context (i.e., $10$ in this case). For example, if a post contains more than one instance of `\textit{feel}' term, we calculate the score individually and consider the maximum one. If the word `\textit{feel}' appears at the $i^{th}$ position in a forum post then the score is calculated as follows:
\begin{equation}
% \scriptsize
Score(+)=\sum_{m=i-k}^{m=i-1}w_m \times f_{+}(SW_m) +  \sum_{n=i+1}^{n=i+k}w_n \times f_{+}(SW_n)
\end{equation}

\begin{equation}
% \scriptsize
Score(-)=\sum_{m=i-k}^{m=i-1}w_m \times f_{-}(SW_m) + \sum_{n=i+1}^{n=i+k}w_n \times f_{-}(SW_n)
\end{equation}
where, $k$ is context window size, weight $w_m=m+k-i+1$ and $w_n=k-n+i+1$.
The aggregate scores $TS_+$ and $TS_-$ of a forum post are calculated as follows:

\begin{equation}
% \scriptsize
\begin{aligned}
TS_+=max(Score_{t=0}(+), Score_{t=1}(+), 
\ldots , Score_{t=T}(+)) 
\end{aligned}
\end{equation}

\begin{equation}
% \scriptsize
\begin{aligned}
TS_-=max(Score_{t=0}(-), Score_{t=1}(-), 
\ldots , Score_{t=T}(-)) 
\end{aligned}
\end{equation}
where, T is the number of sentiment bearing words in the post.
% \end{enumerate}

\subsection{Multi-view Fusion Layer}
We take a multi-view learning approach to combine the various views discussed above into a comprehensive embedding for each medical forum post. We use a  extension of Canonical Correlation Analysis (CCA) \cite{hotelling1992relations} to perform fusion from multiple views. The extended CCA captures maximal information between multiple views and creates a combined representation. The extension of CCA is called the Generalized CCA (GCCA) \cite{carroll1968generalization}, which has been used in the literature to fuse the multiple sources of information into a single source.
GCCA finds $G$,$U_i$ by solving the optimization problem
\begin{equation}
    arg \min_{G,U_i} \sum_i || G-X_iU_i||_{F}^{2}
\end{equation}
such that $G^{T}G = I$.\\
\indent where, $G \in \mathbb{R}^{m \times k}$ contains the fused feature representation matrix, $X_i \in \mathbb{R}^{m \times d_i}$ corresponds to the data matrix for the $i^{th}$ view and $U_i \in \mathbb{R}^{d_i \times k}$ maps from latent space to observable view $i$.
However, since all the views are not equally important, we employ the weighted GCCA (wGCCA) \cite{benton2016learning}. In this representation, we add a weight term to the above equation as follows:
\begin{equation}
    arg \min_{G,U_i} \sum_i w_i|| G-X_iU_i||_{F}^{2}
\end{equation}
such that $G^{T}G = I$ and $w_i \geq 0$ and represents the importance of the $i^{th}$ view in the fusion process. The columns of $G$ are the eigenvectors of $\sum_{i} w_iX_i(X_i^{'}X_i)^{-1}X_i^{'}$, and the solution for $U_i = (X_i^{'}X_i)^{-1}X_i^{'}G$.\\
% We created embeddings using the wGCCA, namely domain adapted embedding, domain adapted with style embedding, and our final eDarkFind embedding as follows. 
\indent We use wGCCA, to obtain the final feature $G$ representation.
%The final classification of the forum post is carried out by a softmax layer having the concatenated task-specific and context features as input. The augmented feature can be represented as follows:
% \begin{dmath}
% \begin{equation}
% % \scriptsize
% \begin{aligned}
% \hat{A}= \underbrace{\hat{d}_{cont}}_{\text{task-agnostic View}} \otimes \underbrace{F_{e}^{f} \otimes  F_{e}^{c}}_{\text{Emotion View}} 
% \otimes \underbrace{F_{s}}_{\text{Sarcasm View}} \otimes \underbrace{F_{P}}_{\text{Personality View}} \otimes\\ \underbrace{WS_{+} \otimes WS_{-} \otimes WS_{O}}_{\text{Word-level Sentiment View}}  \otimes \underbrace{TS_{+}  \otimes  TS_{-}}_{\text{Target-specific Sentiment View
% }}
% \end{aligned}
% \end{equation}
% \end{dmath}
% where $\otimes$ is the concatenation operator.
Finally, the classification of forum post is carried out by the following equation:
\begin{equation}
% \scriptsize
\begin{aligned}
        p(Y=y|M, G) &= softmax_{y}(G^TW+b)\\
        &=\frac{e^{G^TW_y+b_y}}{\sum_{k=1}^{K}e^{G^TW_k+b_k}}
\end{aligned}
\end{equation}
\subsection{Dataset}
We evaluate the performance of the system on the publicly available dataset \cite{yadav2018medical} obtained from popular online health forum\footnote{https://patient.info}. The forum posts were collected from four online discussion groups: \textit{Depression, Allergy, Asthma, and Anxiety}. The dataset consists of $5,621$ medical forum posts related to \textit{Task 1: medical conditions} and $2,734$ forum posts related to the category of \textit{Task 2: medication}. We have extended the previous dataset by including $1$ more class (`Other'), which consider the miscellaneous blog-post. These type of blog post does not explicitly provide any information regarding their medical condition or treatment but are more sort of general enquiry. The more detailed description for each class can be obtained from \cite{yadav2018medical}. 
The detailed description of dataset statistics is provided in Table-\ref{dataset-class1}. 
We perform a 10-fold cross-validation experiment on both the datasets.\\
\textbf{Ethics:} Our project involves analysis of anonymized data that is publicly available and used by the other publication. It does not involve any direct interaction with any individuals or their personally identifiable data. Thus, this study was reviewed by the Wright State University IRB and received an exemption determination.
% \subsection{Training Details}
% We set the hyper-parameters for CNN based content feature network by evaluating the model's performance on the 10-fold cross validation by varying the convolution feature sizes ($100$, $200$, \& $300$). The optimal feature size is turned out to be $300$ with the multiple filter windows of sizes $3$, $4$, and $5$.  
% We use Adam \cite{kingma2014adam} as our optimization method with a learning rate of $0.001$. 
% Training was performed using stochastic gradient descent over mini-batches considering the Adadelta \cite{zeiler2012adadelta} update rule. As a regularizer, we use dropout \cite{hinton2012improving} with a probability of $0.5$. 
\section{Experimental Results}
Here, we present results on the severity assessment task. Thereafter, we will provide technical interpretation of the results followed by ablation study. 
We used Recall, Precision and F\textsubscript{1}-Score to evaluate our proposed task against state-of-the-art relation extractor. As a baseline model, we used \textbf{\textit{BERT}}, \textbf{\textit{BioBERT}}\cite{lee2020biobert}, and multi-task adversarial learning framework \cite{yadav2018multi} to compare our proposed model.\\
\indent We report the performance of our proposed approach along with other baselines in Table \ref{main-results} on task-1 (Medical Condition), and task-2 (Medication). The obtained results shows that BERT model is the best among all the baselines models.  The proposed approach achieves the improvement of $4.56$, $5.17$, and $11.95$ F-Score points for task-1, and $2.76$, $2.81$, and $5.99$  F-Score for task-2 over the baseline $1$, $2$, and $3$ respectively.
Statistical significance test (t-test) shows that improvements over the baselines are significant (\textit{p-value} $<$ $0.05$).\\
\indent To prove the effectiveness of each view, we conduct the ablation experiments on our proposed model. As shown in Table-\ref{feature-ablationl}, the performance of the model shows varying degrees of decline when we remove different view from the model. All the declines are significant with p $\leq$ 0.05 under the t-test.
% In order to investigate the contribution of each feature, we remove one feature at a time from the proposed model and evaluate the system performance. The contributions of different features into the proposed model, is shown in Table \ref{feature-ablationl}.

On the task 1(medical condition), sentiment view seems to be most crucial view, as removal of the word level sentiment drops the F-Score points by $2.91$. Similarly, for the target-specific sentiment, we observe a decline of $1.60$. On the medication task, again sentiment view is found to be important. Removal of word-level and target-specific sentiment view declines the model performance by $3.63$ and $4.30$ F-Score point respectively.
This shows that primarily the sentiment views contribute to determining the health states. The two other significant views affecting the final predictions are emotion (fine) and personality.
As they directly reflect the behavior of a user, the exclusion leads to a decline in the performance. The impact of sarcasm view is found to be smaller as compared to other views, as its removal drops the performance by $0.35$ and $2.75$ F-score points on Medical condition and Medications dataset respectively. Although our analysis shows that the identification of sarcasm is crucial, the little impact could be because of our learning strategy. Since the use of FL in the medical domain is quite different than the general domain.

\section{Discussion}
In this section, we study a couple of cases from both the datasets, where our model correctly identifies various aspects of the health states with the help of NLU views. 
\begin{itemize}
\item \textbf{Case 1: Effect of the emotion view}\\
Consider the following example from task-1:

\textit{``why does anxiety feel like you have to make yourself breathe instead of letting your body breathe on its own. Am'i the only one."}
\\
\indent In the absence of the emotion views, the system misclassifies the post as `Other'. However, the inclusion of the emotion views assists in understanding the users' implicit states of mind, and classifies correctly as `Exist'. The system captures the anger and disgust emotions present in the text, which are highly correlated with the emotion distribution associated with the medical condition category `Exist'. 
\item\textbf{Case 2: Effect of the sarcasm view}\\
In the following post, the user sarcastically expresses his/her condition:\\
\textit{``Lol I'm just a big ball of anxiety fun."}\\
\indent The system misclassifies the post as `Recovered' in the absence of sarcasm view, which may be due to the presence of positive sentiment-bearing words. However, sarcasm view helps the model to predict the class `Exist' correctly. The contextual cues extracted from a post are not always enough to understand someone's feelings and require common sense and background knowledge about the topic of discussion. Such situations are prevalent in the forum posts with very long sentences. 
\item\textbf{Case 3: Effect of the personality view}\\
In our study we find the personality view to be very useful in understanding the users having an anxiety disorder.
Consider the following example from the anxiety group of task-1:\\
\textit{``cutting open my arm. The sensible bit of me says NO just a reaction to my new meds, the other half want to self destruct... I don't know which one is going to win..."}\\
% Here, the system predicted the personality as neuroticism, which is highly correlated with the anxiety. It has a positive relationship with neuroticism and a negative relationship with extraversion. \\
The personality view identified signs of neurotic personality in the above post, which overlaps with the symptoms of anxiety. This helped the system in correctly classifying the user's mental state as having the anxiety disorder.
% captures the user's neuroticism personality, i.e., more likely to experience anxiety. Thus help the system in understanding the user's mental state.
\item \textbf{Case 4: Effect of the sentiment view}\\
As shown in Table \ref{feature-ablationl}, the sentiment views help the model in boosting the performance by nearly 2\% F-Score. In the following example:\\
\textit{``I think it's because I'm afraid of feeling ill when I'm out. This past week I increased citalopram to 20 mg and zI don't know if it's making me feel worse."}\\
The sentiment views capture the negative sentiment-bearing words (i.e., afraid, ill, and worse) and encode this information to assist the model for correctly predicting the class as `Serious Adverse Effect'.
\end{itemize}

\section{Conclusions and Future Work}
 In this paper, we identify the severity of a user's health state by analyzing different medical aspects (such as medical condition and outcome of treatment) from their social media texts. 
 We validate our study by utilizing a benchmark dataset curated from medical web forums. We propose a deep learning model leveraging various NLU views such as emotion, sarcasm, personality, and sentiment along with the textual content for classifying the medical forum posts. The evaluation shows that combining the content view to context views is an effective way to boost the classification performance. In the future, we would like to explore the other facets of a user's health state like `Consequence of a treatment' and `Certainty of a diagnosis'. In addition to sarcasm, we would also like to model other forms of figurative languages like `metaphor' and `irony' which are widely used in social media texts.
\section{Acknowledgement}
Amit Sheth acknowledged partial support from NMH award R01MH105384 “Modeling Social Behavior for Healthcare Utilization in Depression.
All findings and opinions are of authors and not sponsors.
Sriparna Saha would like to acknowledge the support of SERB WOMEN IN EXCELLENCE AWARD 2018. 

\bibliography{bib}
\bibliographystyle{IEEEtran}
\end{document}